\documentclass[review]{elsarticle}

\usepackage{geometry}
\usepackage{amsmath}
\usepackage{graphicx}
\usepackage{array}
\begin{document}

\begin{frontmatter}

\title{Learning to fuse: dynamic integration of multi-source data for accurate battery lifespan prediction}

\author[a]{Shanxuan He}
\author[a]{Zuhong Lin}
\author[a]{Bolun Yu}
\author[b]{Xu Gao}
\author[c]{Biao Long\corref{cor1}}
\author[a]{Jingjing Yao\corref{cor2}}

\affiliation[a]{organization={Center for Environment and Water Resources, College of Chemistry and Chemical Engineering, Central South University}, city={Changsha}, postcode={410083}, country={PR China}}
\affiliation[b]{organization={School of Minerals Processing and Bioengineering, Central South University}, city={Changsha}, postcode={410083}, country={PR China}}
\affiliation[c]{organization={Key Laboratory of Western China's Environmental Systems (Ministry of Education), College of Earth and Environmental Sciences, Lanzhou University}, city={Lanzhou}, state={Gansu}, country={PR China}}

\cortext[cor1]{Corresponding author. E-mail address: longb19@lzu.edu.cn}
\cortext[cor2]{Corresponding author. E-mail address: yaoji0412@163.com}

\begin{abstract}
Accurate prediction of lithium-ion battery lifespan is vital for ensuring operational reliability and reducing maintenance costs in applications like electric vehicles and smart grids. This study presents a hybrid learning framework for precise battery lifespan prediction, integrating dynamic multi-source data fusion with a stacked ensemble (SE) modeling approach. By leveraging heterogeneous datasets from the National Aeronautics and Space Administration (NASA), Center for Advanced Life Cycle Engineering (CALCE), MIT-Stanford-Toyota Research Institute (TRC), and nickel cobalt aluminum (NCA) chemistries, an entropy-based dynamic weighting mechanism mitigates variability across heterogeneous datasets. The SE model combines Ridge regression, long short-term memory (LSTM) networks, and eXtreme Gradient Boosting (XGBoost), effectively capturing temporal dependencies and nonlinear degradation patterns. It achieves a mean absolute error (MAE) of 0.0058, root mean square error (RMSE) of 0.0092, and coefficient of determination ($R^2$) of 0.9839, outperforming established baseline models with a 46.2\% improvement in $R^2$ and an 83.2\% reduction in RMSE. Shapley additive explanations (SHAP) analysis identifies differential discharge capacity (Qdlin) and temperature of measurement (Temp\_m) as critical aging indicators. This scalable, interpretable framework enhances battery health management, supporting optimized maintenance and safety across diverse energy storage systems, thereby contributing to improved battery health management in energy storage systems.
\end{abstract}

\begin{graphicalabstract}
\includegraphics[width=\textwidth]{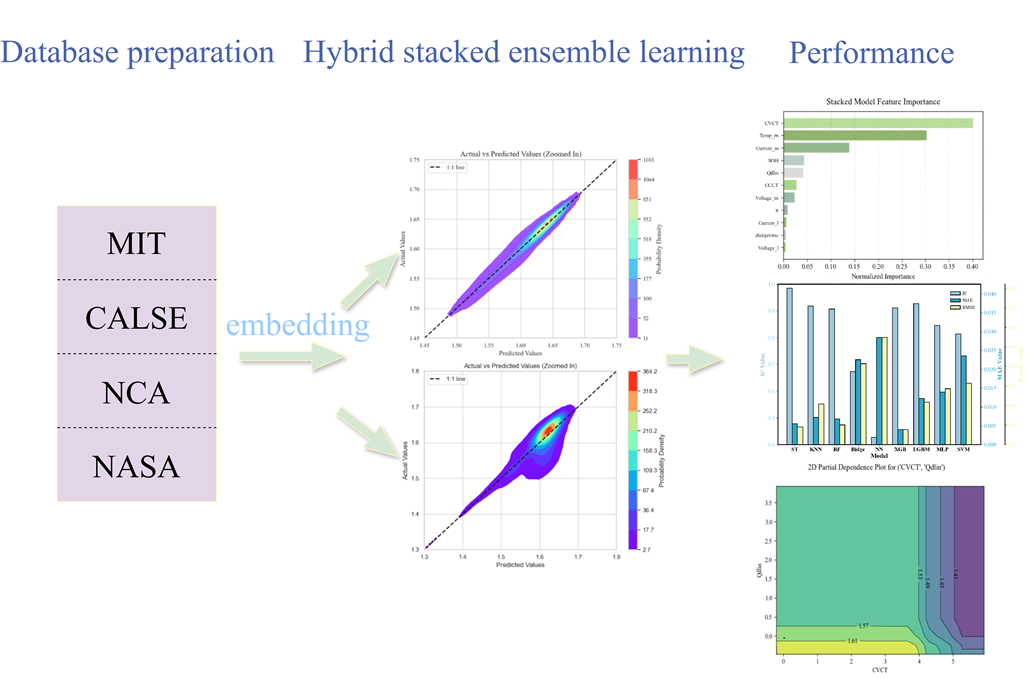}
\end{graphicalabstract}

\begin{highlights}
\item Fuses NASA, CALCE, TRC, NCA datasets to enhance model generalization.
\item Resolves data heterogeneity via dynamic weighting and entropy-based fusion.
\item Hybrid Ridge-LSTM-XGBoost model yields accurate battery lifespan predictions.
\item Feature analysis highlights discharge capacity and temperature as key factors.
\item Outperforms benchmarks and supports practical battery health management.
\end{highlights}

\begin{keyword}
Lithium-ion batteries \sep Battery lifespan prediction \sep Multi-source data fusion \sep Hybrid machine learning \sep Dynamic weighting
\end{keyword}

\end{frontmatter}

\section{Introduction}
\label{sec:intro}

Lithium-ion batteries are essential for electric transportation and smart grid systems. Accurately predicting their lifespan is crucial for ensuring operational reliability, reducing maintenance costs, and promoting sustainable energy practices. The increasing availability of battery cycling data from diverse chemistries provides manufacturers and researchers with an opportunity to develop data-driven models for assessing battery health and predicting performance.

In recent years, machine learning has significantly advanced battery lifespan prediction research. A search of the Web of Science database for "battery prediction" yielded papers published between 2019 and 2024. Analysis of the keywords in these studies revealed that terms such as "machine learning," "state of health" (SOH), and "remaining useful life" were frequently used. This reflects the growing application of machine learning in battery research. Furthermore, a correlation analysis of keywords demonstrates that "machine learning" is applied across diverse domains within battery prediction, reinforcing its status as a highly popular and influential research topic. The abundance of detailed battery data from various types presents researchers and manufacturers with a valuable opportunity to develop models that diagnose and predict battery performance using empirical data. However, the significant variability in battery composition, degradation mechanisms, and real-world operating conditions poses a challenge in creating universally applicable prediction tools. Despite this progress, many existing models depend on single-source datasets, such as National Aeronautics and Space Administration (NASA)'s lithium cobalt oxide data, limiting their adaptability to diverse battery chemistries and dynamic usage scenarios \cite{severson2019data, obi2023briquetting}.

Current machine learning approaches to battery lifespan prediction encounter two key shortcomings. First, their reliance on single-source datasets restricts their effectiveness across varied battery types and real-world conditions \cite{li2024predicting}. Second, traditional algorithms, including support vector machines (SVM), eXtreme Gradient Boosting (XGBoost), and certain deep learning models, struggle to reconcile high-dimensional feature extraction with the modeling of temporal degradation patterns \cite{lin2025rapidly, zhang2022prediction}. For instance, SVMs and XGBoosts excel with low-dimensional data but falter in capturing long-term sequential dependencies \cite{cortes1995support}, whereas sequence-based models like long short-term memory (LSTM) networks demand consistent, densely sampled time-series data—conditions often unmet in heterogeneous battery datasets \cite{yu2019review}. Moreover, the limited interpretability of many sophisticated models impedes their practical deployment in safety-critical contexts \cite{mu2023high}.

To address the limitations in battery lifespan prediction, we propose to develop an integrated machine learning framework that combines dynamic multi-source data fusion with a stacked ensemble approach \cite{yao2023predicting}. We plan to leverage datasets from the National Aeronautics and Space Administration (NASA), the Center for Advanced Life Cycle Engineering (CALCE), MIT-Stanford-TRC, and nickel cobalt aluminum (NCA) chemistries. We intend to implement an entropy-based dynamic weighting mechanism to harmonize these heterogeneous data sources. The proposed predictive model will integrate ridge regression, Long Short-Term Memory (LSTM) networks, and eXtreme Gradient Boosting (XGBoost) within a stacked ensemble (SE) framework, aiming to effectively capture temporal dependencies and nonlinear degradation patterns. This methodology seeks to enhance generalization across diverse battery chemistries and operational conditions while improving model interpretability through Shapley additive explanations (SHAP)-based feature attribution. We anticipate that this framework will surpass established baseline models, delivering a robust and reliable solution for data-driven battery health management.

\section*{Abbreviations}
\label{sec:abbreviations}

\begin{table}[h]
\centering
\begin{tabular}{l l}
\hline
\textbf{Abbreviation} & \textbf{Full name} \\
\hline
CALCE & Center for Advanced Life Cycle Engineering \\
Current\_m & current of measurement \\
CVCT & constant voltage charging time \\
ir & internal resistance \\
KNN & k-nearest neighbors \\
LightGBM & Light Gradient Boosting Machine \\
LSTM & long short-term memory \\
MAE & mean absolute error \\
MIT & Massachusetts Institute of Technology \\
MLP & multi-layer perceptron \\
NASA & National Aeronautics and Space Administration \\
NCA & nickel cobalt aluminum \\
PDP & partial dependence plot \\
Qdlin & differential discharge capacity \\
$R^2$ & coefficient of determination \\
RF & random forest \\
RMSE & root mean square error \\
SE & stacked ensemble \\
SHAP & Shapley additive explanations \\
SOH & state of health \\
SVM & support vector machines \\
Temp\_m & temperature of measurement \\
TRC & Toyota Research Institute \\
Voltage\_m & voltage measure \\
XGBoost & eXtreme gradient boosting \\
\hline
\end{tabular}
\caption{Table of abbreviations}
\label{tab:abbreviations}
\end{table}

\section{Method}
\label{sec:method}

\subsection{Database integration and preprocessing}
\label{subsec:database}

\begin{figure}[ht]
\centering
\includegraphics[width=0.8\textwidth]{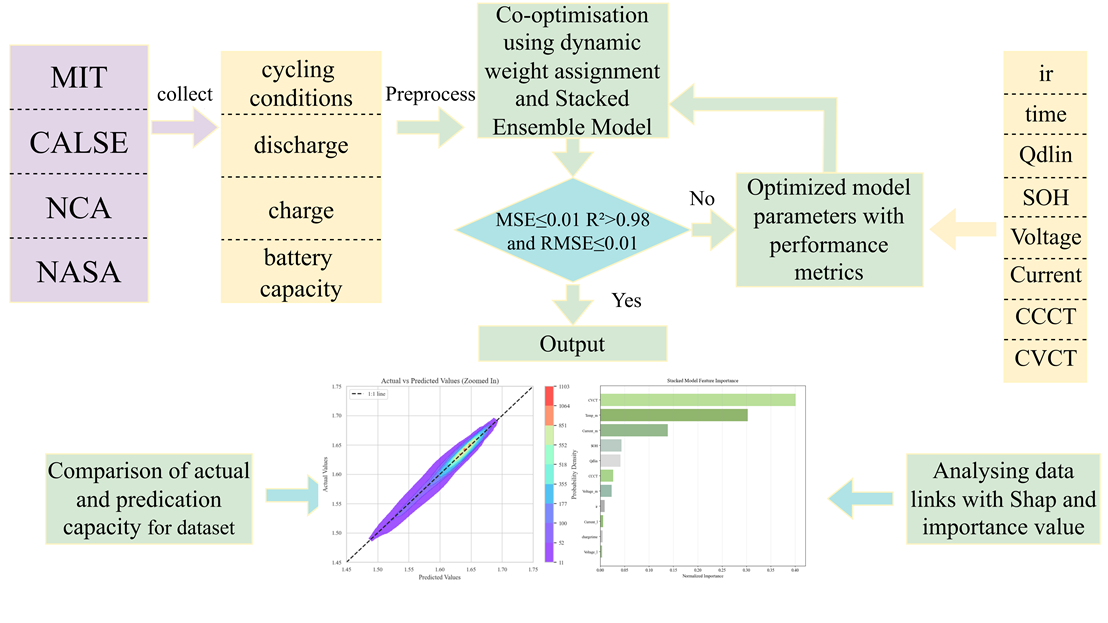}
\caption{Stacked ensemble model flowchart.}
\label{fig:fig1}
\end{figure}

The workflow for lithium-ion battery lifespan prediction, shown in Fig. \ref{fig:fig1}, includes multi-source dataset integration, preprocessing, model optimization, and analysis. The first step in the workflow is the collection of a comprehensive database. Multiple datasets were utilized to ensure a robust analysis of battery performance, sourced from publicly accessible lithium-ion battery datasets \cite{dos2021lithium}. These include NCA battery data derived from voltage relaxation analyses \cite{zhu2022data}, four aging datasets (B0005, B0006, B0007, and B0018) from the NASA \cite{khumprom2019data}, four operational datasets from a collaboration between the Massachusetts Institute of Technology (MIT) and the Toyota Research Institute (TRC) \cite{severson2019data}, and a cycling dataset covering cells CS2-35 to CS2-38 from the CALCE \cite{yun2020state}. This multi-source framework enables comprehensive validation across diverse battery chemistries, operational conditions, and degradation mechanisms.

\begin{figure}[ht]
\centering
\includegraphics[width=0.8\textwidth]{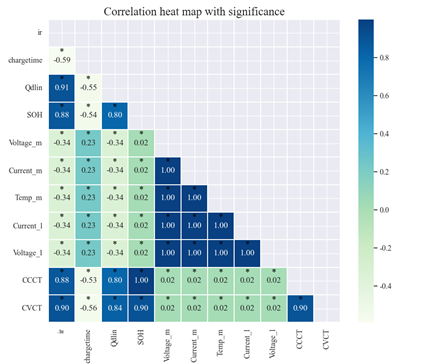}
\caption{Correlation heat map with significance.}
\label{fig:fig2}
\end{figure}

The preprocessing workflow begins with data aggregation, consolidating heterogeneous datasets into a unified table. Features such as charge/discharge profiles are transformed into numerical representations using embedding methods to preserve intrinsic correlations \cite{tang2022early}. Feature selection is then performed by calculating the Pearson Correlation Coefficient for all feature pairs, visualized through a Correlation Heat Map with Significance, as shown in Fig. \ref{fig:fig2}. The heatmap displays Pearson coefficients ranging from -0.59 to 1.00 for the 11 features retained after eliminating multicollinearity \cite{paulson2022feature, goldstein2015peeking}. To address cross-dataset challenges, such as dimensionality mismatches and feature misalignment, an entropy-based dynamic weighting mechanism is implemented, ensuring effective integration of multi-source data. This mechanism employs a dual-criteria evaluation system that dynamically adjusts model weights based on both prediction accuracy and uncertainty quantification. Specifically, each base model's contribution weight $w_i$ is determined by

\begin{equation}
w_i = \frac{R_i^2}{1 + \operatorname{Var}\left(y_{\text{pred}}^i\right)}
\label{eq:weight}
\end{equation}

where $R_i^2$ denotes cross-validated determination coefficient and $\operatorname{Var}\left(y_{\text{pred}}^i\right)$ represents prediction variance as entropy proxy. The final weights are normalized to form a convex combination, ensuring stable numerical properties. This optimized preprocessing approach results in a streamlined dataset with 11 key parameters, facilitating accurate and robust battery lifespan prediction.

\subsection{Model algorithms}
\label{subsec:algorithms}

Several machine learning algorithms were evaluated to assess their effectiveness in predicting battery performance. We evaluated a variety of machine learning algorithms, including ensemble methods such as XGBoost \cite{chen2016xgboost}, Light Gradient Boosting Machine (LightGBM) \cite{ke2017lightgbm}, and random forest (RF) \cite{biau2016random}. Additionally, we assessed other techniques like support vector regression (SVR) \cite{cortes1995support}, multi-layer perceptron (MLP) \cite{hornik1989multilayer}, Keras neural network \cite{haghighat2021sciann}, k-nearest neighbors (KNN) \cite{cover1967nearest}, and ridge regression \cite{hoerl2000ridge}. This selection enabled a thorough exploration of battery data characteristics across diverse paradigms. Ensemble methods bolstered robustness by integrating weak learners \cite{dietterich2000ensemble}, while SVR and neural networks adeptly handled high-dimensional, nonlinear data. KNN and Ridge Regression provided insights into local patterns and linear relationships, respectively. Hyperparameters were tuned using Bayesian optimization \cite{chen2016xgboost}, and performance was evaluated via root mean square error (RMSE), mean absolute error (MAE), coefficient of determination ($R^2$), and 5-fold cross-validation.

A SE model was constructed to enhance prediction accuracy and robustness. This framework integrated ridge regression \cite{hoerl2000ridge, wolpert1992stacked}, XGBoost \cite{chen2016xgboost}, and LSTM as base models, capitalizing on their complementary strengths \cite{yu2019review, mu2023high}. Ridge regression captured linear relationships, XGBoost addressed nonlinearities and feature interactions \cite{lin2024enhancing}, and LSTM extracted temporal patterns from charge-discharge cycles. A meta-model, based on ridge regression, fused predictions from the base models, effectively combining linear, nonlinear, and temporal features.

\subsection{Model interpretation}
\label{subsec:interpretation}

To enhance the interpretability of our hybrid learning framework and gain insights into the model's behavior and predictions, we employ three key interpretation methods—residual plot analysis, feature importance analysis using SHAP \cite{lundberg2020local, lin2024unveiling}, and 2D partial dependence plot (PDP) analysis \cite{goldstein2015peeking, friedman2001greedy, greenwell2017pdp}. These approaches collectively provide a robust understanding of the model's performance and decision-making process. These methods enhance transparency and enable practical application in predicting battery lifespan under complex, real-world conditions.

Residual plot analysis involves examining the residuals, which represent the differences between predicted and actual battery discharge capacities. By plotting these residuals, we can evaluate the model's accuracy and detect any systematic errors or patterns that might suggest limitations in capturing the data's underlying dynamics. An ideal outcome shows residuals randomly distributed around zero, indicating that the model effectively accounts for the degradation patterns present in the multi-source battery datasets. Feature importance analysis using SHAP quantifies the contribution of each feature to the model's predictions, offering a clear view of what drives the outcomes. Based on cooperative game theory, SHAP values allow us to interpret individual predictions and identify critical features. This method enhances transparency, making it particularly valuable for safety-critical applications where understanding the model's reasoning is essential. 2D PDP analysis visualizes the relationship between one or two features and the predicted outcome, averaging out the effects of other variables. This technique reveals how specific features or their interactions influence battery capacity predictions. Smooth, continuous contours in the 2D PDP, as visualized through heatmaps or contour plots, suggest effective modeling of nonlinear feature interactions. Abrupt discontinuities or irregular patterns, by contrast, may highlight regions where the model struggles to generalize, offering actionable insights into both its predictive strengths and limitations for subsequent refinement.

\section{Results and Discussion}
\label{sec:results}

\subsection{Model establishment and screening process}
\label{subsec:model_screening}

The establishment and screening of predictive models for lithium-ion battery lifespan involve a systematic evaluation of multiple machine learning algorithms, culminating in the development of a robust SE model. The process begins with the integration of diverse datasets from NASA \cite{khumprom2019data}, CALCE \cite{yun2020state}, MIT-Stanford-TRC \cite{severson2019data}, and NCA chemistries \cite{zhu2022data}, followed by meticulous preprocessing to ensure data consistency. We applied thorough preprocessing steps, like choosing features based on the Pearson Correlation Coefficient, to blend different types of data together and tackle issues with variables overlapping too much. We evaluated a range of machine learning algorithms to determine their effectiveness, including ridge regression \cite{hoerl2000ridge}, XGBoost \cite{chen2016xgboost}, LightGBM \cite{ke2017lightgbm}, RF \cite{biau2016random}, support vector regression (SVR) \cite{cortes1995support}, MLP \cite{hornik1989multilayer}, Keras neural network (NN) \cite{haghighat2021sciann}, KNN \cite{cover1967nearest}, and our own SE model \cite{wolpert1992stacked}. The SE model, which integrates Ridge Regression for linear relationships, XGBoost for nonlinear relationships, and LSTM networks for capturing temporal dependencies, demonstrated superior performance. Its superior performance is attributed to the effective integration of the strengths of each base model.

\begin{figure}[ht]
\centering
\includegraphics[width=0.72\textwidth]{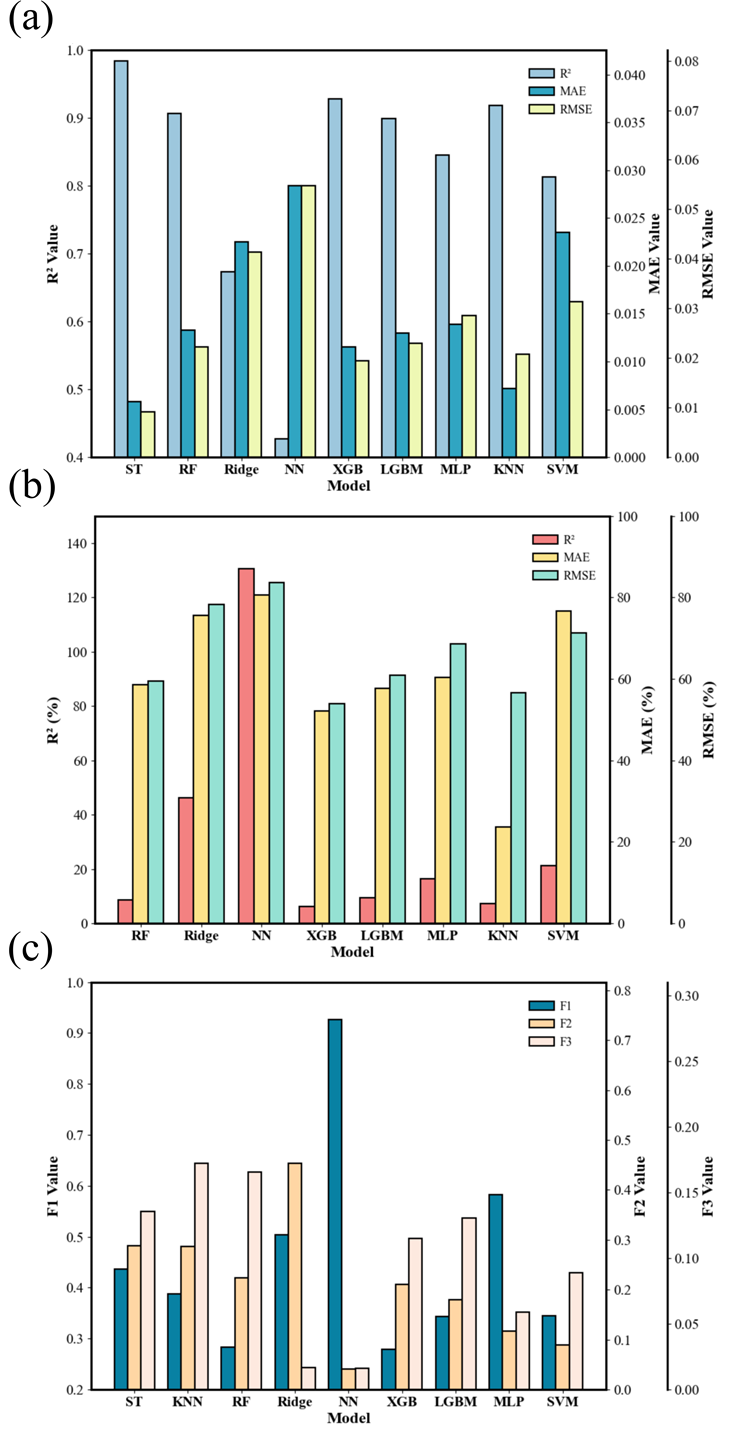}
\caption{Comparative analysis of regression model performance charts.}
\label{fig:fig3}
\end{figure}

We looked at how good each model was by using key measures like $R^2$, MAE, and RMSE. Fig. \ref{fig:fig3} provides a bar graph comparing these metrics across all models. The SE model achieved an $R^2$ of 0.9839, MAE of 0.0058, and RMSE of 0.0092, indicating high predictive accuracy. This tells us it's great at picking up the complicated ways batteries wear down. In contrast, Ridge Regression achieved a lower $R^2$ of 0.6731, MAE of 0.0225, and RMSE of 0.0414, while the Keras Neural Network demonstrated even poorer performance, achieving an $R^2$ of 0.4266, an MAE of 0.0284, and an RMSE of 0.0548. Other solid models, like XGBoost ($R^2=0.9277$), LightGBM ($R^2=0.8991$), and KNN ($R^2=0.918$), put up good numbers but didn’t quite reach the SE model’s level of accuracy.

Fig. \ref{fig:fig3} breaks down how much better the SE model is compared to the usual approaches, showing the improvements in percentages. Against ridge regression, it bumps up $R^2$ by 46.2\%, slashes MAE by 74.2\%, and drops RMSE by 77.8\%. When you put it next to the Keras Neural Network, it lifts $R^2$ by a huge 130.7\%, cuts MAE by 79.6\%, and shrinks RMSE by 83.2\%. Even compared to strong performers like XGBoost, the SE model pushes $R^2$ up by 6.1\%, trims MAE by 49.6\%, and lowers RMSE by 52.8\%. These results demonstrate the model’s robustness and ability to generalize across diverse datasets. For further analysis, Fig. \ref{fig:fig3} presents a bar chart illustrating the feature importance for each model, providing insights into the key factors influencing their predictive accuracy. Feature importance is determined using SHAP \cite{lundberg2020local}, a method rooted in cooperative game theory that quantifies each feature’s contribution to model predictions. The x-axis lists the models (SE, KNN, RF, Ridge, NN, XGBoost, LightGBM, MLP, SVR), while the y-axis quantifies the importance of three key features: differential discharge capacity (Qdlin), constant voltage charging time (CVCT), and temperature measure (Temp\_m), color-coded in blue, orange, and pink, respectively. These features consistently rank among the most influential across models, highlighting their critical role in battery performance prediction.

For the top-performing SE model ($R^2=0.9839$), feature importance values are notably high for CVCT (0.4363), Temp\_m (0.2881), and current of measurement (Current\_m) (0.1358). This prioritization reflects the model’s adept use of charging dynamics and operational conditions, contributing to its exceptional accuracy \cite{roman2021machine}. Similarly, XGBoost ($R^2=0.9277$) assigns significant importance to CVCT (0.2792), Qdlin (0.2107), and SOH (0.1168), while KNN ($R^2=0.918$) emphasizes Qdlin (0.3874), SOH (0.2863), and internal resistance (ir) (0.0164). These models effectively balance electrochemical and operational features, enhancing their predictive capabilities \cite{shu2021state}.

Conversely, lower-performing models exhibit distinct patterns. Ridge regression ($R^2=0.6731$) assigns a disproportionately high importance to voltage measure (Voltage\_m) (0.5043) and Temp\_m (0.4535), yet its predictive accuracy suffers due to its inability to model nonlinear relationships effectively \cite{goldstein2015peeking, shu2021state}. Similarly, the Keras Neural Network ($R^2=0.4266$) prioritizes Voltage\_m (0.0890) and Temp\_m but struggles with overall performance, likely due to insufficient capture of complex degradation dynamics. These discrepancies underscore the critical interplay between algorithm selection and feature prioritization. Ensemble and tree-based models like SE and XGBoost excel by leveraging features such as Qdlin, CVCT, and Temp\_m, which align with the physical mechanisms of battery degradation, whereas linear or less flexible models like ridge regression falter despite strong feature emphasis.

The combined insights affirm the SE model as the optimal choice for battery lifespan prediction. Its hybrid architecture, refined through hyperparameter tuning with Optuna and interpretability analysis via SHAP and 2D PDPs \cite{goldstein2015peeking, friedman2001greedy, greenwell2017pdp}, delivers a scalable and transparent framework. These findings have significant implications for practical battery health management, enabling precise lifespan predictions and informed operational strategies.

\subsection{Residual plot comparison}
\label{subsec:residual}

\begin{figure}[ht]
\centering
\includegraphics[width=0.8\textwidth]{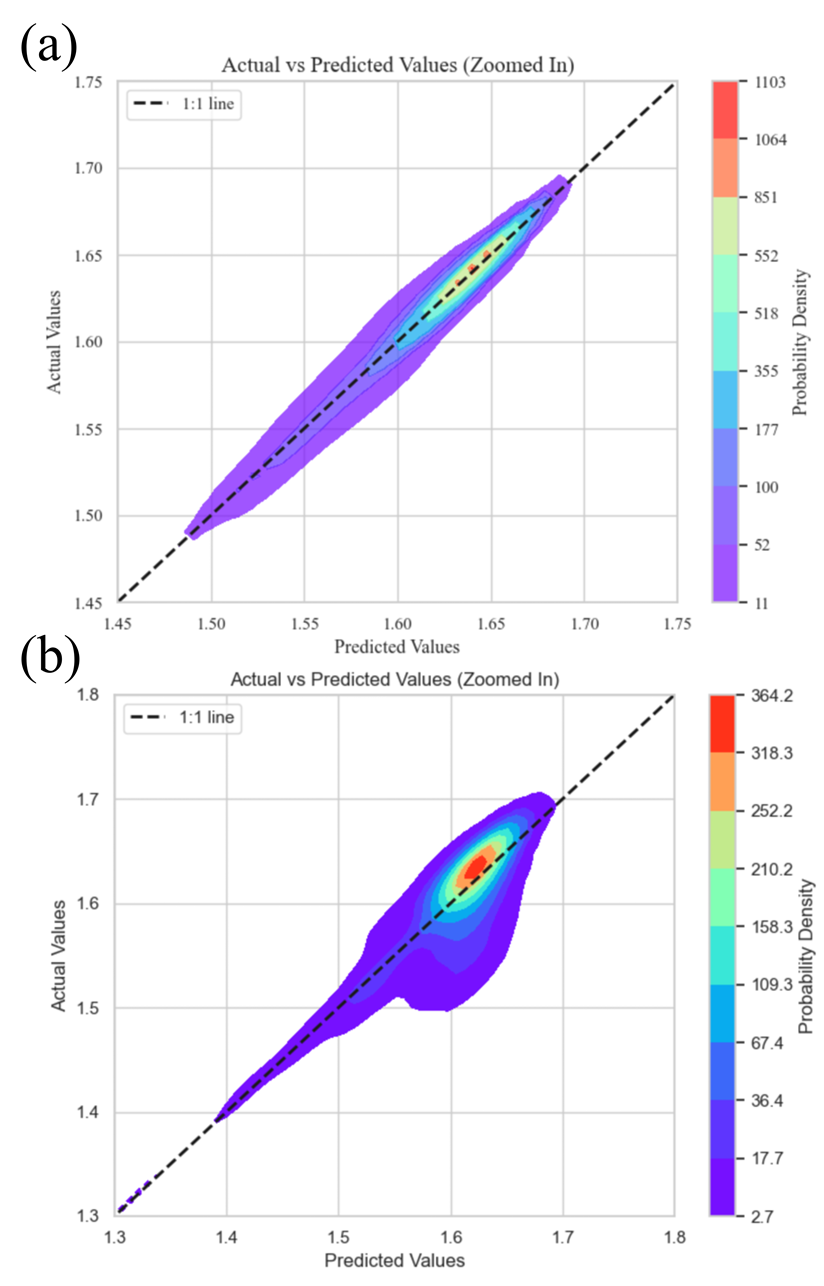}
\caption{Comparison of residual plots. (a) The SE model (b) The XGBoost model.}
\label{fig:fig4}
\end{figure}

Residual plots provide critical insights into model calibration and robustness by visualizing the differences between predicted and actual battery discharge capacities. In this study, comparative residual density plots were analyzed for the SE and XGBoost. For the SE model (Fig. \ref{fig:fig4}a), residuals exhibited a tight concentration along the ideal 1:1 diagonal line, with high-density regions (yellow to red) that indicate high predictive accuracy and consistency across mid- and high-capacity regimes. This narrow spread and close alignment confirm the model’s precision, as the residuals are randomly distributed around zero, which suggests the absence of systematic bias \cite{eleftheriadis2024comparative, ma2022residual}. In contrast, the XGBoost model (Fig. \ref{fig:fig4}b) exhibits broader dispersion around the 1:1 line, with high-density regions (red) concentrated at 1.5-1.6 Ah but extending further into the 1.7-1.8 Ah range, reflecting greater variability and reduced accuracy compared to SE. The pronounced divergence in residual patterns highlights the SE model’s superior performance in capturing complex battery capacity dynamics.

\subsection{Feature importance analysis}
\label{subsec:feature_importance}

Feature importance was assessed using SHAP values, providing a transparent quantification of each feature’s contribution to the model’s predictions \cite{lundberg2020local}. Across all tested algorithms, three features consistently emerged as the most influential: Qdlin, CVCT, Temp\_m. These features align with key physical descriptors of lithium-ion battery ageing: Qdlin reflects capacity fade, CVCT indicates electrochemical stress during charging, and Temp\_m captures thermally induced degradation mechanisms \cite{strobl2008conditional, krupp2021incremental}.

High-performing models, such as the SE ($R^2=0.9839$), XGBoost ($R^2=0.9277$), and KNN ($R^2=0.918$), effectively leveraged these features for predictive inference. The SE, in particular, placed significant emphasis on differential discharge capacity (Qdlin), which reflects its generalizability across diverse datasets and battery chemistries, a finding that is consistent with prior studies on capacity-related predictors in battery health monitoring. In contrast, conventional models like ridge regression ($R^2=0.6731$) and MLP neural networks ($R^2=0.4266$) exhibited a more diffuse feature importance distribution, with weaker alignment to physically relevant ageing indicators \cite{hornik1989multilayer}. This highlights the limitations of linear and shallow nonlinear models in capturing high-order interactions in heterogeneous multi-source data.

These results underscore two key implications: first, feature selection informed by physical interpretability (e.g., Qdlin as a proxy for active material loss) enhances model robustness; second, architectures with dynamic feature prioritization, such as ensemble and tree-based models, excel in generalizing across varied usage scenarios \cite{strobl2008conditional}. The prominence of Qdlin, CVCT, and Temp\_m suggests their potential as core indicators for simplified surrogate models or low-latency BMS implementations. This interpretability, supported by SHAP-based analysis, is crucial for transparent and safe deployment in battery health management, aligning with advances in explainable AI for energy systems \cite{zhao2024artificial}.

\begin{figure}[ht]
\centering
\includegraphics[width=0.8\textwidth]{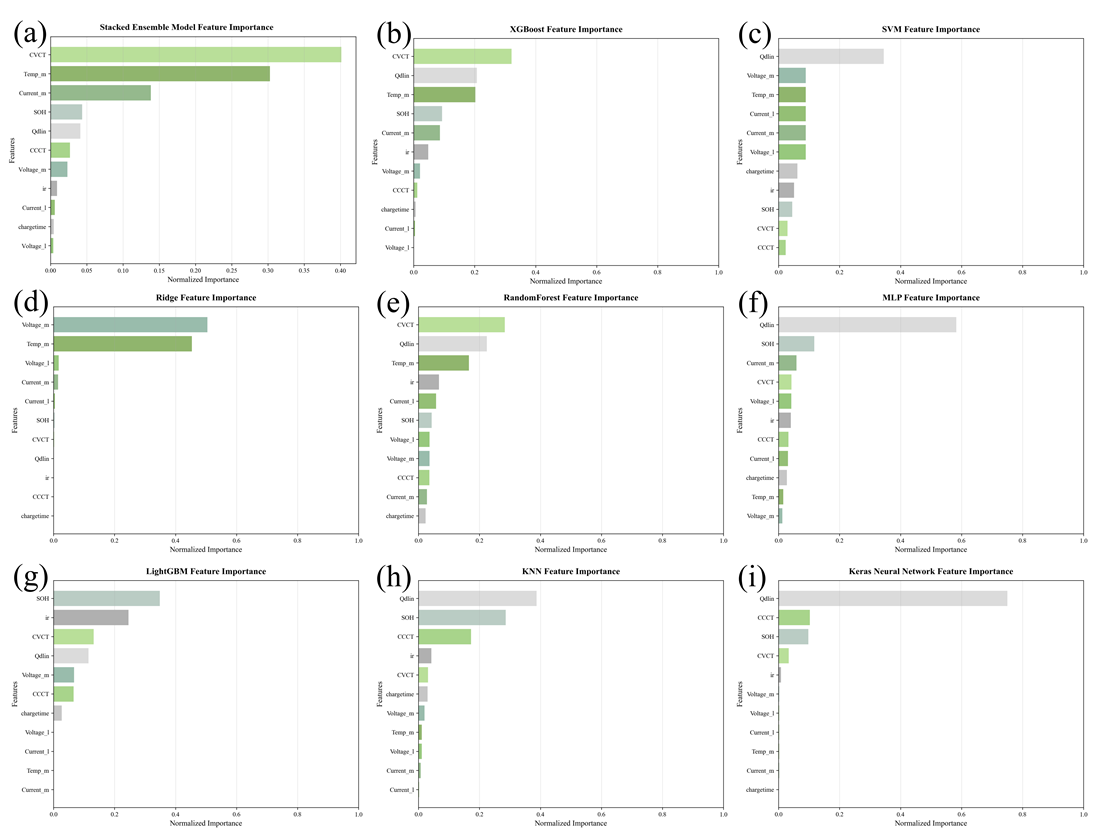}
\caption{Feature importance analysis chart. This composite figure integrates nine individual bar charts, each illustrating the feature importance analysis for a distinct machine learning model: (a) Stacked Ensemble, (b) XGBoost, (c) SVM, (d) Ridge, (e) RF, (f) MLP, (g) LightGBM, (h) KNN, and (i) Keras Neural Network.}
\label{fig:fig5}
\end{figure}

\subsection{2D Partial dependence plot Analysis}
\label{subsec:pdp}

\begin{figure}[ht]
\centering
\includegraphics[width=0.8\textwidth]{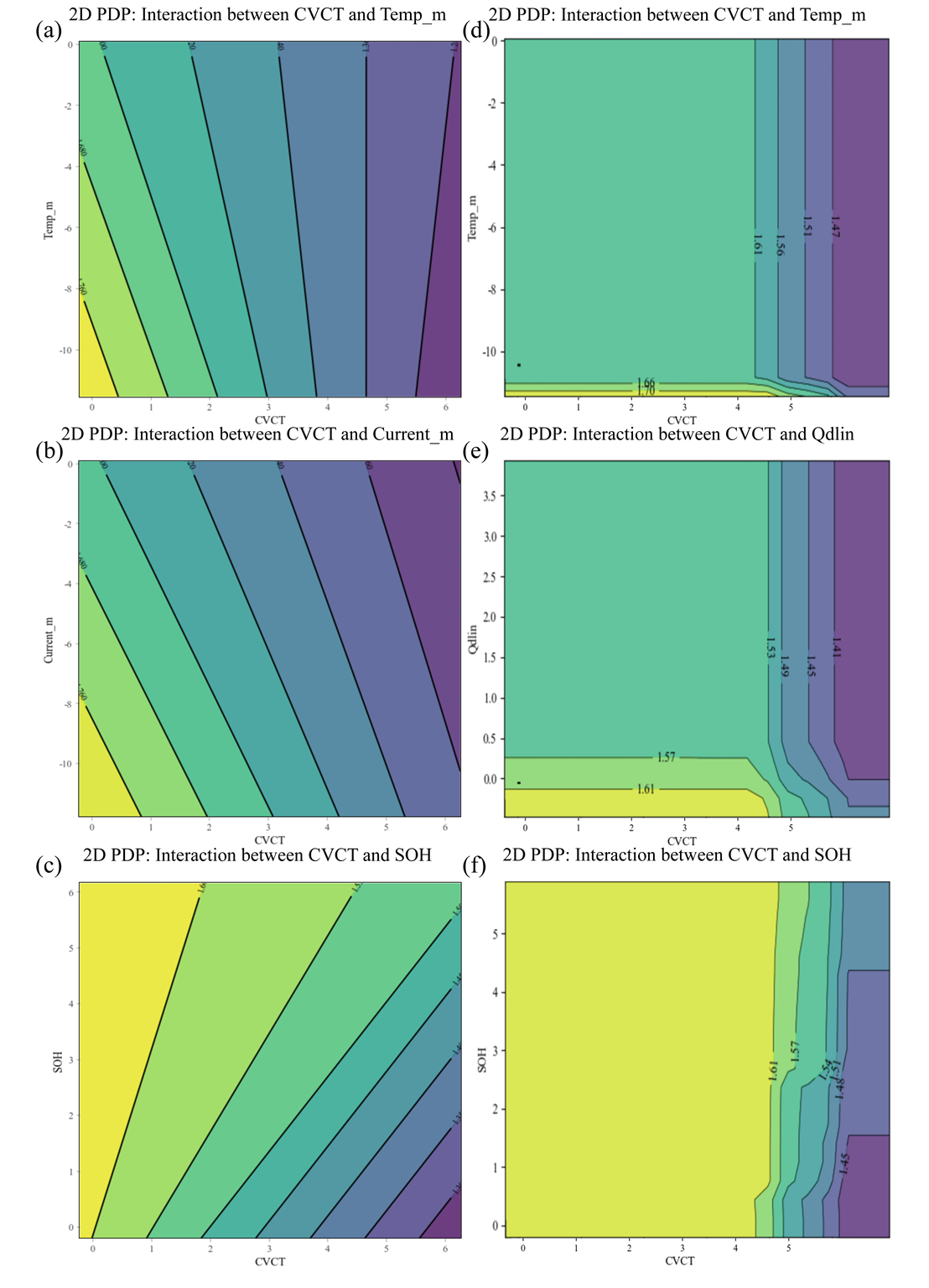}
\caption{2D PDP: Stacked Ensemble vs. XGBoost. Panels (a-c) represent the SE model. Panels (d-f) represent the XGBoost model}
\label{fig:fig6}
\end{figure}

2D PDPs were used to examine the interactions between key features and their effects on predicted battery discharge capacity \cite{paulson2022feature, goldstein2015peeking, hastie2009elements}. For the SE model (Fig. \ref{fig:fig6}a-c), the PDP surfaces displayed smooth, continuous, and well-structured contours, indicating effective modeling of nonlinear feature interactions in a physically coherent manner. For instance, in the CVCT-SOH plot (Fig. \ref{fig:fig6}c), predicted capacity rises steadily from approximately 1.25 Ah to 1.65 Ah as both variables increase, demonstrating a smooth progression devoid of abrupt transitions. Similar monotonic and diagonal trends in the CVCT-Temp\_m and CVCT-Current\_m panels (Fig. \ref{fig:fig6}a-b) demonstrate the model’s ability to synthesize correlated inputs seamlessly. This smoothness reflects the hybrid integration of ridge regression (linearity), XGBoost (nonlinearity), and LSTM (temporal dependencies), producing a harmonized output surface.

Conversely, XGBoost’s PDPs (Fig. \ref{fig:fig6}d-f) exhibited discontinuous, step-like transitions and boxy regions, suggesting limitations in capturing smooth feature interactions. These artifacts, typical of decision-tree-based learners under sparse or misaligned data distributions, indicate unstable gradients and reduced generalizability \cite{wolpert1992stacked}. The irregular contours highlight XGBoost’s challenges in modeling the nuanced, continuous relationships critical to battery degradation prediction.

The SE’s smooth PDP surfaces enhance both predictive accuracy and interpretability, making it suitable for applications like sensitivity analysis, control parameter tuning, and automated policy learning in BMS \cite{dietterich2000ensemble}. In safety-critical environments, the absence of abrupt contour changes prevents erroneous outputs that could trigger false alarms, a key advantage over less stable models, as supported by studies emphasizing the importance of smooth state estimation and model stability. These findings demonstrate that the hybrid model, through its stacked generalization approach \cite{wolpert1992stacked}, not only achieves numerical superiority but also structurally improves the learning and explanation of degradation dynamics, bridging data-driven predictions with domain-specific reasoning, as evidenced by joint estimation techniques \cite{beelen2021joint}.

Extending the analysis to additional models reveals a broader spectrum of predictive capabilities for battery discharge capacity. The Ridge model produces smooth, linear contour trends, as exemplified by the predictable linear growth in Voltage\_m vs. Current\_m, demonstrating its strength in capturing continuous feature interactions with minimal risk of overfitting. This robustness stems from its linear regression foundation, making it a stable meta-learner; however, its linear assumption struggles to reflect the nonlinear dynamics inherent in battery degradation, limiting its effectiveness in complex prediction scenarios. In contrast, the RF model exhibits discontinuous, box-like contours, such as the abrupt capacity decline in CVCT vs. ir, indicative of its decision-tree-based structure’s difficulty in modeling smooth, gradual feature interactions \cite{friedman2001greedy}. This segmentation suggests reduced generalization ability in dynamic battery systems, where smooth transitions are critical. Similarly, the LightGBM model displays discontinuous, blocky patterns, like the sudden shift in SOH vs. Qdlin, highlighting its limitations in smooth interaction modeling, particularly in sparse data regions, which constrains its ability to capture the progressive changes central to battery degradation.

The MLP model offers smooth, nonlinear profiles, such as the multimodal capacity peak in Qdlin vs. SOH, hinting at its potential to capture complex interactions, yet its mediocre fit points to optimization challenges, possibly due to suboptimal feature selection or tuning. Meanwhile, the Keras Neural Network leans toward linear trends, as seen in the diagonal capacity increase in Qdlin vs. Current, revealing a failure to capture essential nonlinear dynamics, likely due to an overly simplistic architecture or training approach, rendering it less suited for intricate battery prediction tasks. In contrast, the KNN model exhibits smooth nonlinear contours, as exemplified by the complex peak observed in the Qdlin versus ir relationship, demonstrating its capability to capture intricate interactions. This adaptability positions KNN as a strong contender for battery capacity prediction, rivaling the SE’s prowess. Collectively, these insights highlight a range of model strengths: the SE and KNN excel in nonlinear modeling with smooth, interpretable PDP surfaces, while Random Forest and LightGBM falter in continuity due to their tree-based nature \cite{eleftheriadis2024comparative}. Ridge provides stability in simpler, linear contexts, whereas Keras underperforms in complex scenarios, and MLP’s potential remains hampered by optimization needs, affirming the SE’s superior balance of accuracy and interpretability in battery management systems \cite{liu2022interpretable}.

\section{Conclusion}
\label{sec:conclusion}

In this study, we introduce a robust hybrid learning framework for accurately predicting lithium-ion battery lifespan across diverse chemistries and operational conditions. By integrating multi-source datasets from NASA, CALCE, MIT-Stanford-TRC, and NCA chemistries through an entropy-based dynamic weighting mechanism, our approach effectively mitigates data heterogeneity and variability inherent in real-world battery applications. The framework employs a SE model combining ridge regression, LSTM networks, and XGBoost, harnessing their complementary strengths to capture linear relationships, temporal dependencies, and nonlinear degradation patterns. Experimental results demonstrate exceptional predictive performance, achieving a MAE of 0.0058, RMSE of 0.0092, and $R^2$ of 0.9839, surpassing baseline models with a 46.2\% improvement in $R^2$ and an 83.2\% reduction in RMSE. SHAP-based feature analysis highlights Qdlin and Temp\_m as critical aging predictors, yielding physically interpretable insights into degradation mechanisms. Additionally, 2D PDP analysis reveals that the SE model produces smooth, continuous contour surfaces, effectively capturing nonlinear feature interactions—unlike models such as XGBoost and RF, which exhibit discontinuous patterns. This smoothness enhances both predictive accuracy and interpretability, providing practical benefits for real-world battery health management applications. With its strong generalizability across diverse datasets and transparency through interpretable analysis, this framework serves as a scalable tool for data-driven battery health management, delivering practical value in real-world applications.

\clearpage

\setcounter{section}{0}

\renewcommand{\appendixname}{Supporting Information}

\renewcommand{\thefigure}{S\arabic{figure}}
\renewcommand{\thetable}{S\arabic{table}}
\renewcommand{\thepage}{S\arabic{page}}
\setcounter{page}{1}
\setcounter{figure}{0}
\setcounter{table}{0}
\renewcommand{\theequation}{S\arabic{equation}}  
\setcounter{equation}{0}
\addtocontents{toc}{\protect\setcounter{tocdepth}{1}}

\part*{Supporting Information}

\begin{table}[ht]
\centering
\caption{Detailed information of datasets used in the study}
\label{tab:S1}
\begin{tabular}{p{4cm} p{2cm} p{2cm} p{4cm} >{\centering\arraybackslash}p{1.5cm}}
\hline
Primary Applications & Battery Type & Quantity/ Range & Test Conditions & Reference \\
\hline
Capacity estimation, machine learning model development & NCA, NCM, NCA+NCM & 130 cells & Various cycling protocols, voltage relaxation curve characteristics & [11] \\
Lifetime prediction, aging analysis & Lithium-ion & B0005$\sim$B0018 & Charge/discharge/impedance tests, multi-temperature conditions, deep discharge aging & [12] \\
Cycle life prediction, early-stage data modeling & Lithium-ion & 124 cells & Hundreds to thousands of cycles, voltage/current/temperature measurements & [1] \\
State estimation, remaining useful life prediction, reliability analysis & LiCoO$_2$ batteries & CS2-35 to CS2-38 & Room temperature, cycling/storage/dynamic profiles, impedance measurements & [13] \\
\hline
\end{tabular}
\end{table}

\begin{table}[ht]
\centering
\caption{Importance values of the top three features for each model}
\label{tab:S2}
\small
\begin{tabular}{c c c c c c c c c c}
\hline
Model & ST & RF & Ridge & NN & XGB & LGBM & MLP & KNN & SVM \\
\hline
Feature 1 & CVCT & CVCT & Voltage\_m & Qdlin & CVCT & SOH & Qdlin & Qdlin & Qdlin \\
 & 0.4363 & 0.2832 & 0.5043 & 0.9269 & 0.2792 & 0.3438 & 0.5827 & 0.3874 & 0.3450 \\
Feature 2 & Temp\_m & Qdlin & Temp\_m & chargetime & Qdlin & ir & SOH & SOH & Voltage\_m \\
 & 0.2881 & 0.2242 & 0.4535 & 0.0408 & 0.2107 & 0.1802 & 0.1168 & 0.2863 & 0.0890 \\
Feature 3 & Current\_m & Temp\_m & Voltage\_1 & ir & Temp\_m & Qdlin & Current\_m & CCCT & Temp\_m \\
 & 0.1358 & 0.1654 & 0.0168 & 0.0164 & 0.1553 & 0.1396 & 0.0589 & 0.1724 & 0.0890 \\
\hline
\end{tabular}
\end{table}

\begin{table}[ht]
\centering
\caption{Regression evaluation metrics for each model}
\label{tab:S3}
\begin{tabular}{cccccccccc}
\hline
 & ST & RF & Ridge & NN & XGB & LGBM & MLP & KNN & SVM \\
\hline
R2 & 0.9839 & 0.9060 & 0.6731 & 0.4266 & 0.9277 & 0.8991 & 0.8445 & 0.918 & 0.8123 \\
MAE & 0.0058 & 0.0133 & 0.0225 & 0.0284 & 0.0115 & 0.013 & 0.0139 & 0.0072 & 0.0235 \\
RMSE & 0.0092 & 0.0222 & 0.0414 & 0.0548 & 0.0195 & 0.023 & 0.0286 & 0.0207 & 0.0314 \\
\hline
\end{tabular}
\end{table}

\begin{figure}[ht]
\centering
\includegraphics[width=0.8\textwidth]{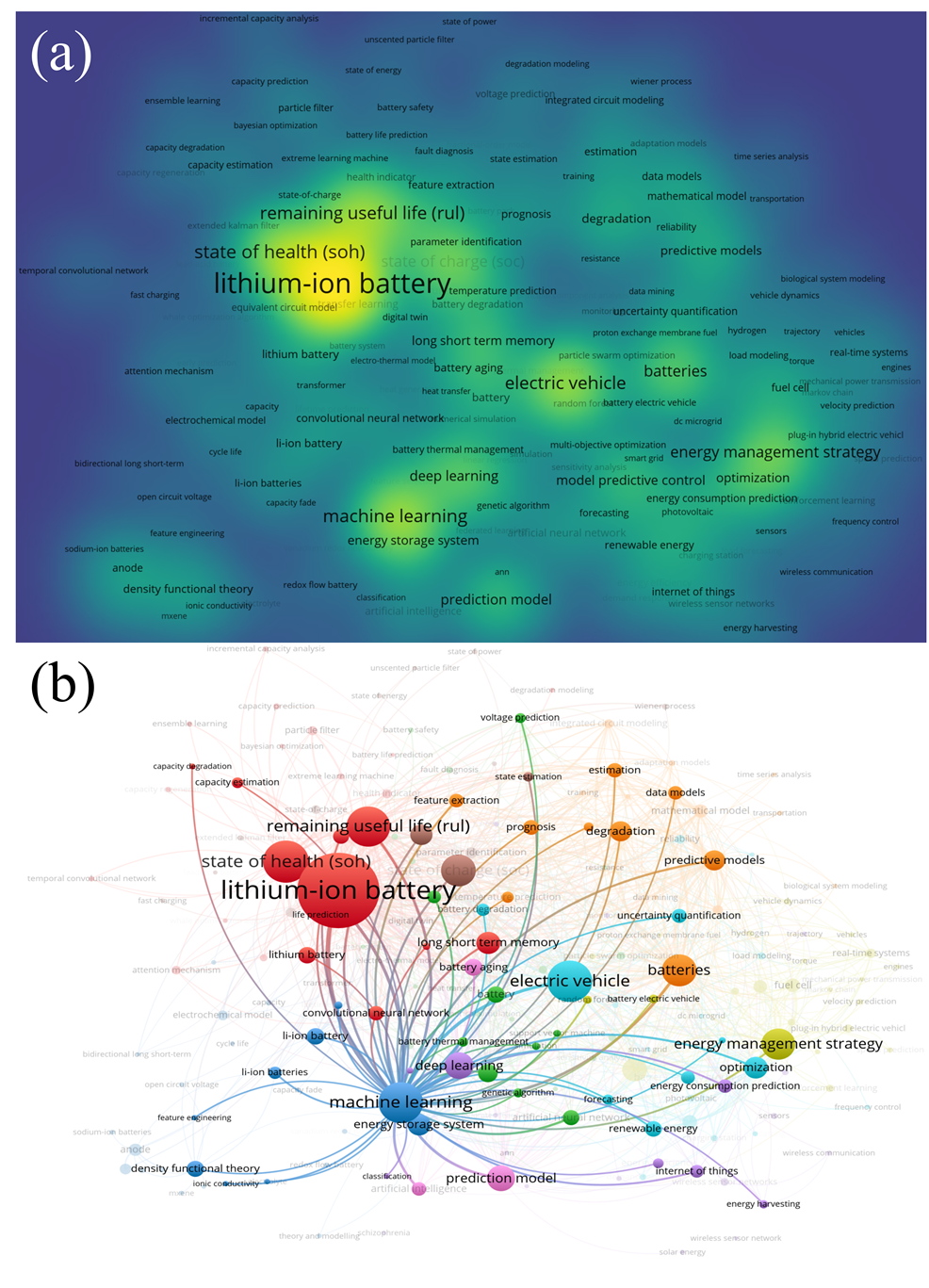}
\caption{Bibliometric Analysis of Battery Prediction Research. (a) Frequency of keywords in battery prediction research. (b) Correlations among common keywords.}
\label{fig:S1}
\end{figure}

\begin{figure}[ht]
\centering
\includegraphics[width=0.4\textwidth]{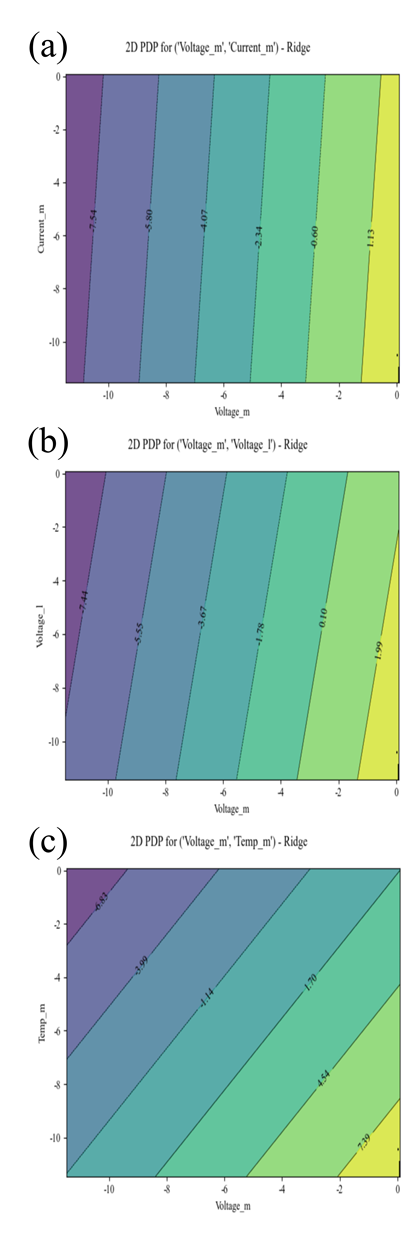}
\caption{2D PDP plots for the Ridge Model. (a) Smooth linear trend between Voltage\_m and Current\_m. (b) Smooth linear trend between Voltage\_m and Voltage\_l. (c) Smooth linear trend between Voltage\_m and Temp\_m.}
\label{fig:S2}
\end{figure}

\begin{figure}[ht]
\centering
\includegraphics[width=0.4\textwidth]{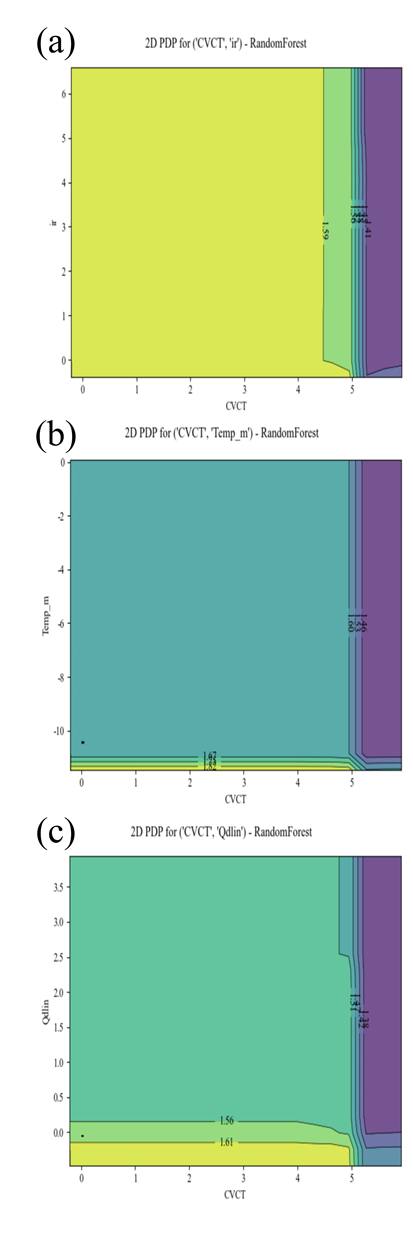}
\caption{2D PDP plots for Random Forest. (a) Discontinuous pattern between CVCT and ir. (b) Minimal variation between CVCT and Temp\_m. (c) Sharp transition between CVCT and Qdlin.}
\label{fig:S3}
\end{figure}

\begin{figure}[ht]
\centering
\includegraphics[width=0.4\textwidth]{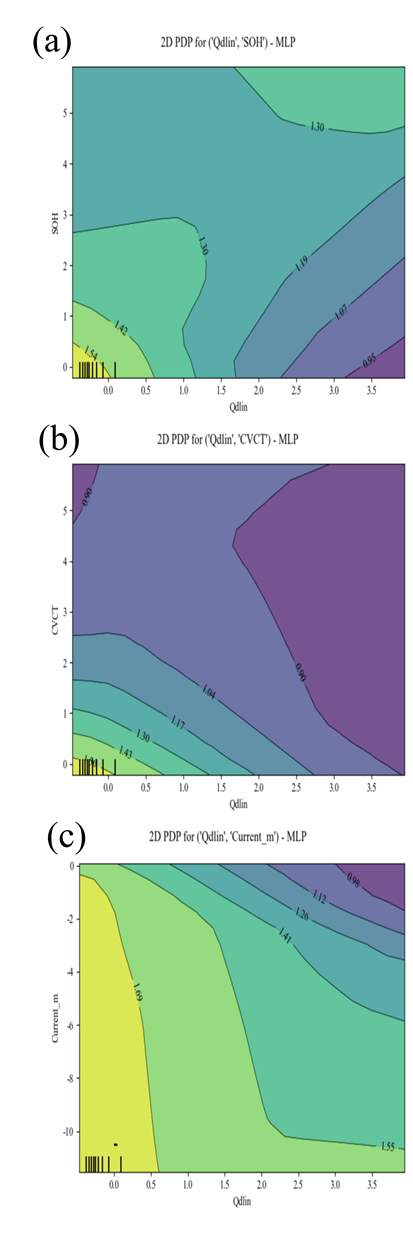}
\caption{2D PDP plots for MLP. (a) Non-linear interaction between Qdlin and SOH. (b) Smooth non-linear contours between Qdlin and CVCT. (c) Mixed linear and non-linear trends between Qdlin and Current\_m.}
\label{fig:S4}
\end{figure}

\begin{figure}[ht]
\centering
\includegraphics[width=0.4\textwidth]{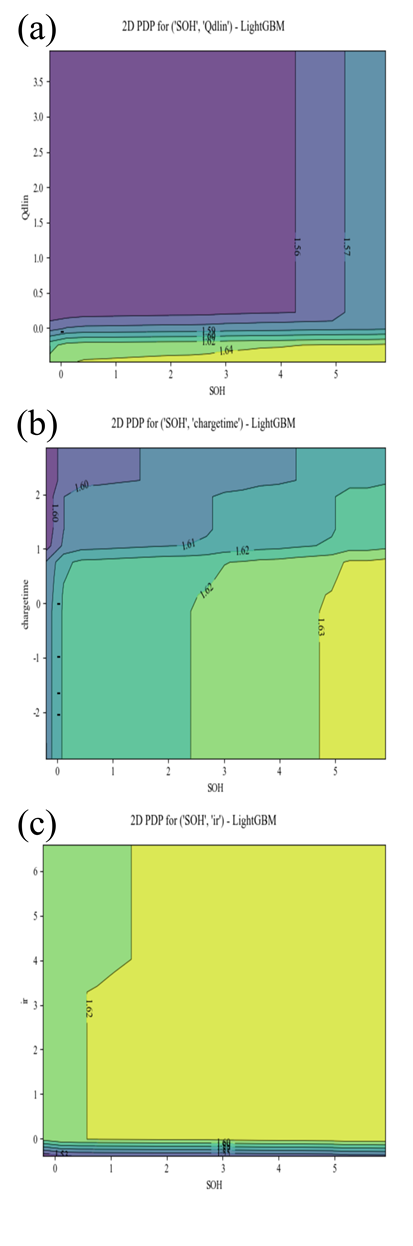}
\caption{2D PDP plots for LightGBM. (a) Sharp transition between SOH and Qdlin. (b) Stepped contours between SOH and chargetime. (c) Minimal variation between SOH and Ir.}
\label{fig:S5}
\end{figure}

\begin{figure}[ht]
\centering
\includegraphics[width=0.4\textwidth]{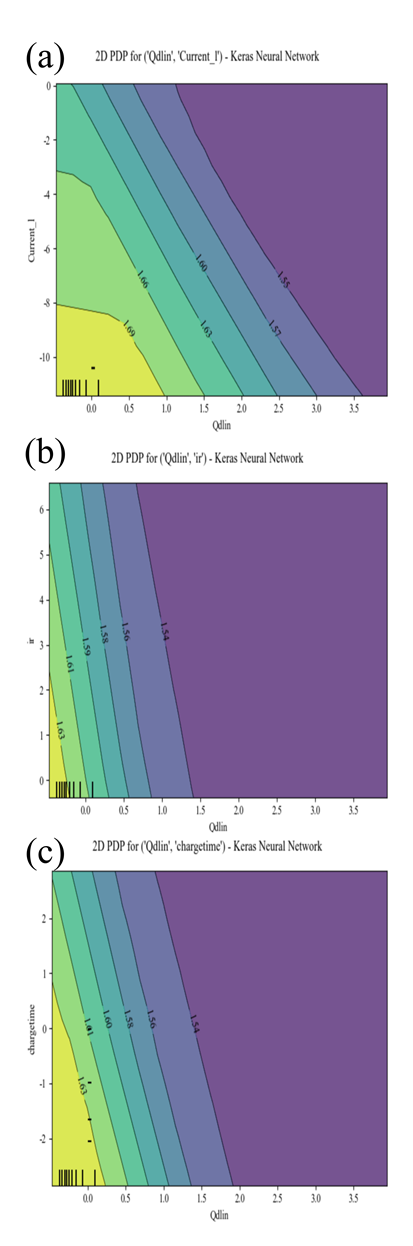}
\caption{2D PDP plots for Keras Neural Network. (a) Linear trend between Qdlin and Current\_m. (b) Linear trend between Qdlin and ir. (c) Linear trend between Qdlin and chargetime.}
\label{fig:S6}
\end{figure}

\begin{figure}[ht]
\centering
\includegraphics[width=0.4\textwidth]{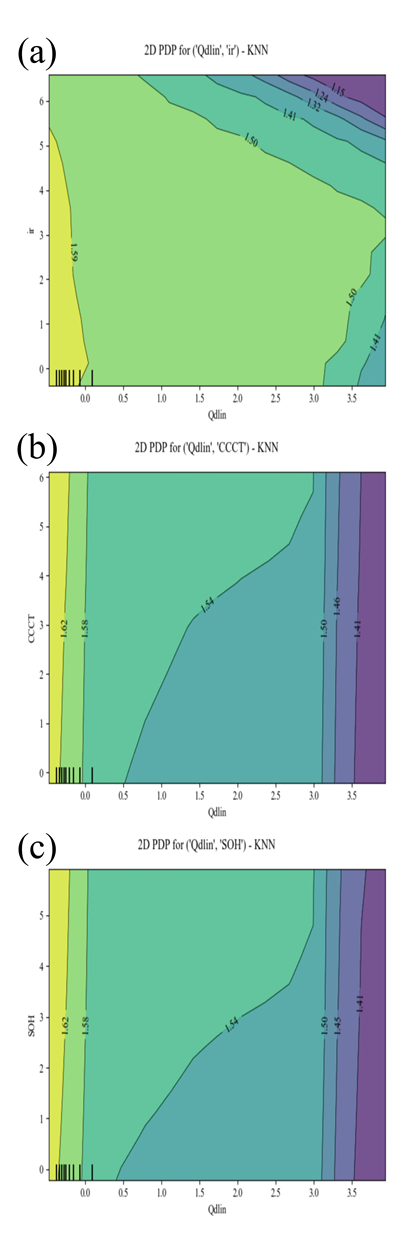}
\caption{2D PDP plots for KNN. (a) Non-linear trend between Qdlin and ir. (b) Non-linear trend between Qdlin and CCCT. (c) Non-linear trend between Qdlin and SOH.}
\label{fig:S7}
\end{figure}

\end{document}